\documentclass{article}
\usepackage{spconf,amsmath,amssymb,graphicx,booktabs,multirow,tikz,url}
\newcommand{\result}[2]{#1$\mathord{\pm}$#2}

\title{PROGRESSIVE-VIEW ON-POLICY DISTILLATION FOR REGIONAL-TO-GLOBAL TRANSFER IN MULTIMODAL LLMS}

\name{Shanfeng Huang, Zhou Fang, Song Xiao, Hai Du}
\address{Baidu Inc.\\
         \{huangshanfeng, fangzhou07, xiaosong, duhai02\}@baidu.com}

\begin{document}
\ninept
\maketitle

\begin{abstract}
Regional-to-global distillation uses crop-conditioned guidance to improve full-image understanding.
The challenge is to effectively transfer the teacher's crop-based advantage to the student's full-image inference.
We propose progressive-view on-policy distillation (PVD), which shifts the student's view distribution
from the crop toward the full image through an intermediate aspect-preserving padded crop. The padded
crop preserves regional content while matching the full image's visual-token grid. Across stages, the
view mixture assigns increasing probability to the full image. A lightweight regional-advantage weighting
reallocates token-level supervision using the crop-conditioned teacher--student log-probability gap.
Evaluated under each sampled input, it applies mild reweighting when the gap is small and emphasizes higher-gap tokens when the gap widens.
A Jensen--Shannon metric decomposition interprets this schedule as a shift from matched-input
imitation toward the deployment objective. Across benchmarks spanning perception, visual mathematics and general
multimodal question answering, PVD-full reaches an average accuracy of
$77.51$ over three seeds, improving on the reward-free distillation baseline by $2.01$ points and on
its reward-matched variant by $1.00$ point. In the reward-free setting, PVD-distill still gains $1.16$
points.
\end{abstract}

\begin{keywords}
Multimodal large language models, on-policy self-distillation, fine-grained visual understanding, regional-to-global transfer, progressive view sampling
\end{keywords}

\section{INTRODUCTION}
\label{sec:intro}

Fine-grained visual detail remains a bottleneck for multimodal large language
models, because the evidence a question depends on often occupies a tiny fraction of a
high-resolution image. Much prior work resolves this at inference time through guided
visual search, tree-structured zooming, or agentic crop-and-reask loops~\cite{vstar,zoomeye},
which recover the detail at the cost of added latency and deployment machinery for every query.

Regional-to-global distillation transfers this benefit into the
weights~\cite{visionopd,zoombench}. Conditioning the same model on the annotated region makes it a \emph{privileged teacher}
that sees the same evidence at a larger effective scale and produces sharper next-token distributions.
These distributions supervise the full-image student on its own rollouts~\cite{gkd,tmlopd}.
Inference uses one student forward pass on the full image.
On-policy distillation avoids the exposure bias of offline supervision~\cite{gkd,minillm,tmlopd} and
has been extended to self-distillation with the same model serving both roles~\cite{visionopd,uopsd}.
Token-level variants show that teacher--student disagreement concentrates in a small subset of response
tokens, motivating learnability-based token selection or visual-advantage-based
reweighting~\cite{taopd,vaopd}. In regional-to-global distillation, however, the standard objective
uses the student's full-image view from the outset and weights response tokens uniformly, even though
the privileged-view advantage is concentrated in a minority of tokens.

We address these problems with \textbf{progressive-view on-policy distillation} (PVD), which integrates
progressive view sampling with view-dependent token weighting in a single objective. The student's view
distribution shifts in stages from the crop toward the full image through an aspect-preserving padded
crop, and every stage keeps mass on more than one view. The padded crop preserves regional content on
the full image's visual-token grid, separating token-grid alignment from the later introduction of
surrounding context. Progressive sampling therefore bridges the input mismatch over graded stages, and
view-dependent weighting concentrates supervision on the tokens the privileged view resolves. The two
components are linked through the teacher--student gap under the sampled view. RA remains mild when
this gap is small and becomes more selective for high-gap tokens in later, more deployment-aligned
view mixtures.
Our contributions are:
\begin{itemize}
\itemsep0pt
  \item We propose PVD, a \emph{progressive} regional-to-global transfer objective that shifts
  sampling mass in stages from privileged crops toward the deployment view. View-conditioned
  regional-advantage (RA) token weighting is a complementary refinement that reuses the same
  teacher--student gap and adds no extra forward pass.
  \item An exposure-matched control isolates the source of the gain: the progressive schedule beats a
  static mixture with the same expected exposure to each view, so temporal progression, not added
  crop exposure, accounts for most of the reward-free improvement over Vision-OPD.
  \item A Jensen--Shannon metric decomposition (Section~\ref{ssec:sched}) explains why the schedule
  runs in this direction: the crop aligns the teacher and student visual inputs, while the full image
  matches the deployment input.
  \item Augmenting the curriculum with a complementary task-reward term, PVD-full reaches $77.51$,
  $+2.01$ over the reward-free Vision-OPD and $+1.00$ over a reward-matched control, and neither
  signal alone matches the combination.
\end{itemize}

\begin{figure}[t]
\centering
\includegraphics[width=\columnwidth]{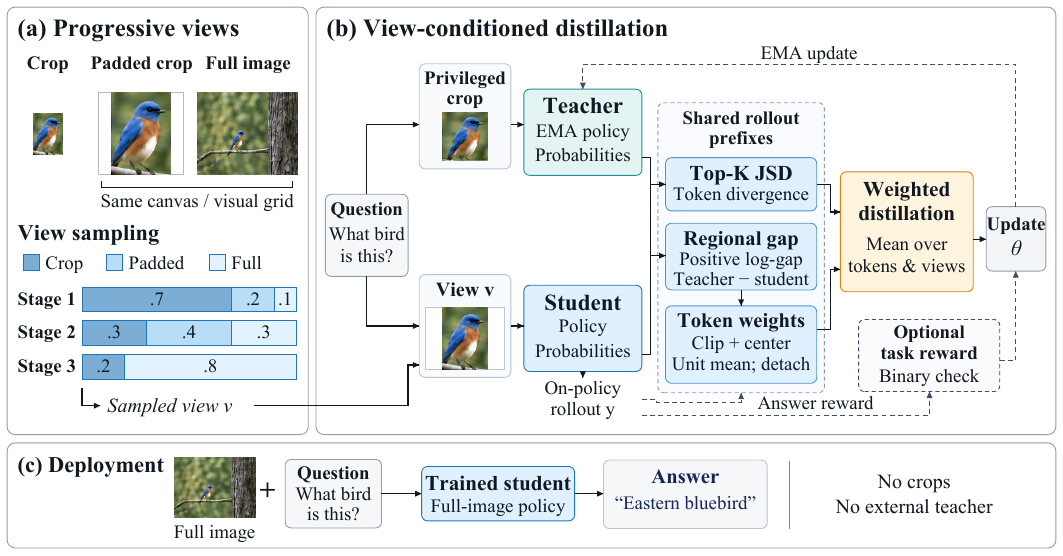}
\caption{Progressive-view on-policy distillation. Student training uses stagewise mixtures of a
regional crop, a grid-aligned padded crop containing the same regional content, and the full image,
with sampling mass shifting toward the deployment view while the teacher remains conditioned on the
privileged crop. View-conditioned teacher--student matching produces the top-$K$ JSD and
regional-advantage (RA) token weights. At deployment, only the student processes the full image.}
\label{fig:method}
\end{figure}

\section{PROPOSED METHOD}
\label{sec:method}

\subsection{Regional-to-global self-distillation}

Let $x$ be an image, $q$ a question and $b$ a region annotation marking the area the question
depends on. We use a view $v$ to denote one of the student input constructions. The student is
$\pi_\theta$ conditioned on $v$, and the teacher
$\pi_{T}$ is an exponential-moving average of the same weights, conditioned on the crop $x_b$. Rollouts are drawn on-policy under the student's own
view, $y \sim \pi_\theta(\cdot \mid v, q)$, and the per-token loss is a top-$K$ Jensen--Shannon
divergence between the two next-token distributions,
\begin{equation}
\ell_t(v) = \mathrm{JSD}_K\!\big(\pi_{T}(\cdot \mid x_b, q, y_{<t}) \,\big\|\,
                                 \pi_\theta(\cdot \mid v, q, y_{<t})\big).
\label{eq:jsd}
\end{equation}
The support is the student's top-$K$ vocabulary entries, where the teacher's log-probabilities are
gathered, and both are renormalized with a tail bucket for the residual mass.

\subsection{A view-conditioned transfer objective}
\label{ssec:obj}

Vision-OPD~\cite{visionopd}, the representative method in this direction, fixes $v = x$ and averages
$\ell_t$ over the response mask, giving every token the same weight. PVD replaces this fixed-view
uniform average with a single objective in which the student's observation is itself a random
variable, and the discrepancy that observation induces decides where supervision is spent:
\begin{gather}
\mathcal{L}_{\mathrm{PVD}} = \mathbb{E}_{v \sim p(\tau),\, y \sim \pi_\theta(\cdot \mid v, q)}\Big[\tfrac{1}{|y|}\textstyle\sum_t w_t(v)\,\ell_t(v)\Big],
\label{eq:pvd}\\[1pt]
g_t(v) = \big[\log \pi_{T}(y_t \mid x_b, q, y_{<t}) - \log \pi_\theta(y_t \mid v, q, y_{<t})\big]_{+},
\label{eq:gap}\\[1pt]
\bar{g}_t(v) = \min\!\big(g_t(v),\, g_{\max}\big) - \big\langle \min(g(v),\, g_{\max})\big\rangle_t,
\label{eq:centered-gap}\\[1pt]
w_t(v) = \mathcal{N}\big[\mathrm{clip}\big(1 + \lambda\, \bar{g}_t(v),\; w_{\min},\; w_{\max}\big)\big],
\label{eq:w}
\end{gather}
Here $p(\tau)$ is the view distribution at progress $\tau = s/S$, $\langle\cdot\rangle_t$ averages over
the response mask, and the \emph{privileged-view gap} $g_t(v)$ is treated as constant in the backward
pass. The operator $\mathcal{N}$ normalizes the weights to unit mean, letting $\lambda$ redistribute
supervision without changing its total. The weighting reuses log-probabilities that
\eqref{eq:jsd} already computes and requires no additional box signal, auxiliary scorer or forward
pass beyond the crop-conditioned teacher. We refer to it as regional-advantage (RA) weighting, since it
scales each token's supervision by its privileged-view gap $g_t$.

The two dimensions are coupled because $w_t(v)$ is evaluated under the \emph{sampled} view. When the
privileged-view gap is small, the weighting is less selective. As $p(\tau)$ shifts probability mass
toward later, more deployment-aligned views, larger gaps concentrate transfer on tokens with a greater
teacher advantage. Progressive view sampling therefore
shapes token-level supervision throughout training, and the ablation in Section~\ref{ssec:ablation} isolates the gain of the
schedule.

We impose two conditions on the schedule. Order the views $v_1, \dots, v_M$ from the teacher's crop to the
deployment image, $v_1 = x_b$ and $v_M = x$, and let
$p(\tau)=(p_1(\tau),\dots,p_M(\tau))$ be piecewise constant on $R$
stages with boundaries $0 = \tau_0 < \dots < \tau_R = 1$. \emph{Monotone progression}:
$\sum_{k \le j} p_k(\tau)$ is non-increasing in $\tau$ for every $j < M$, so mass moves toward later
views. \emph{Terminal alignment}: $p_M$ dominates the last stage, so the final phase samples and scores
predominantly under the deployment view. The teacher's conditioning is never scheduled. We use
$M = 3$ (crop, aspect-preserving padded crop, full image) and $R = 3$. A direct crop-to-full-image
transition changes both token-grid geometry and visual content, including the region's relative scale.
We therefore insert the \emph{aspect-preserving, white-canvas padded crop} as an intermediate view.
Given source-image dimensions $(W,H)$ and crop dimensions $(W_b,H_b)$, we resize the crop by
$\min(W/W_b,H/H_b)$ with aspect ratio preserved (using Lanczos interpolation), and paste the result
at the center of a $(W,H)$ white canvas. The resized crop therefore spans the canvas in at least one
dimension, and any remaining margins are white, so no pixels outside the crop are retained. The crop
and padded crop contain the same regional content, while the padded crop and full image share the same
canvas dimensions and visual-token grid. The intermediate view therefore aligns token-grid geometry
while preserving crop-only content. The full-image view then introduces the surrounding context and
restores the region's relative scale at deployment.
Fig.~\ref{fig:method} illustrates how PVD works, from the progressive view schedule and
view-conditioned teacher--student matching to deployment on the full image.

\subsection{A metric interpretation of PVD}
\label{ssec:sched}

Training only at the fixed deployment view, as Vision-OPD does, exposes the student to the full
crop-to-image discrepancy from the outset. The annotated region covers a small fraction of the image, so after the vision
encoder's resizing much of the evidence the teacher conditions on is severely under-resolved in the student's input.
For a full-distribution metric view of \eqref{eq:jsd}, let $\pi^{v}_{\phi}$ denote the next-token
distribution and $d(\cdot,\cdot) = \mathrm{JSD}(\cdot\,\|\,\cdot)^{1/2}$, which is a metric on
distributions~\cite{jsmetric}. For any student view $v$ the triangle inequality gives
\begin{equation}
\underbrace{d\big(\pi^{x_b}_{T},\, \pi^{x}_{\theta}\big)}_{\text{deployment discrepancy}}
\;\le\;
\underbrace{d\big(\pi^{x_b}_{T},\, \pi^{v}_{\theta}\big)}_{\text{metric surrogate}}
\;+\;
\underbrace{d\big(\pi^{v}_{\theta},\, \pi^{x}_{\theta}\big)}_{\delta_X(v):\ \text{deployment slack}} ,
\label{eq:bound}
\end{equation}
To identify how the view changes this surrogate, define two student-side distances that vanish at
opposite ends of the view order:
\begin{equation}
\delta_T(v) = d\big(\pi^{x_b}_{\theta},\, \pi^{v}_{\theta}\big), \qquad
\delta_X(v) = d\big(\pi^{v}_{\theta},\, \pi^{x}_{\theta}\big).
\label{eq:deltas}
\end{equation}
The \emph{crop-to-view distance} $\delta_T$ measures how far the student's distribution moves as its
input construction shifts from the teacher's crop to the sampled view. It captures changes in both
token-grid geometry and visual content through their effect on the student's distribution and serves
as a proxy for surrogate difficulty. The
\emph{deployment slack} $\delta_X$ is the gap between what
is optimized and what is deployed. At $v = x_b$, visual inputs match and $\delta_T = 0$, so the
metric surrogate contains only the EMA parameter mismatch while deployment slack remains. At $v = x$,
$\delta_X = 0$ and the bound is tight: the metric surrogate equals the deployment discrepancy and
includes the full crop-to-image discrepancy. The crop endpoint aligns the
visual inputs, whereas the full-image endpoint aligns training with deployment.

The decomposition thus exposes a trade-off between matching the teacher's visual input and matching
the deployment input. For fixed parameters and a common response prefix, if the candidate views
satisfy $\delta_X(v_1)\ge\dots\ge\delta_X(v_M)=0$, shifting probability mass toward later views
reduces the expected deployment slack, while a larger $\delta_T$ typically makes the surrogate harder
and can induce more selective token weighting. This motivates a schedule that begins with substantial crop exposure
and ends predominantly on the deployment view, shifting training from matched-input imitation toward
the deployment objective. The decomposition provides an interpretation of the schedule, and
Section~\ref{sec:exp} evaluates it empirically.

The token allocation admits an exact decomposition. For a fixed sampled response, let
$\langle\cdot\rangle_t$ average over its valid tokens. Since $\langle w\rangle_t=1$,
\begin{equation}
\langle w\ell\rangle_t
=\langle\ell\rangle_t+\operatorname{Cov}_t(w,\ell).
\label{eq:allocation}
\end{equation}
Indeed, expanding $\langle(w-1)(\ell-\langle\ell\rangle_t)\rangle_t$ gives the second term.
When weight clipping is inactive, $w_t=1+\lambda\bar g_t$, so this correction is
$\lambda\operatorname{Cov}_t(\bar g,\ell)$. Thus RA redistributes a unit-mean supervision budget,
emphasizing divergence on tokens where the privileged-view teacher holds a larger advantage over
the student. Progressive view sampling determines the inputs and rollouts on which this allocation
operates. Setting $\lambda=0$ retains the progressive objective with uniform weights, and additionally
fixing $v=x$ recovers Vision-OPD.

\subsection{Joint training with task reward}

PVD supervises the student toward the teacher's distribution without checking whether its final
answer is correct, while a binary task reward checks only the answer and says nothing about which
tokens to change. The two signals therefore carry different information. We combine them in a single
per-token objective,
\begin{equation}
\mathcal{L} = \mathcal{L}_{\mathrm{PVD}} + \beta\, \mathcal{L}_{\mathrm{PG}},
\label{eq:joint}
\end{equation}
where $\mathcal{L}_{\mathrm{PG}}$ is a group-relative policy-gradient loss~\cite{grpo} on
the binary answer reward. Both terms apply to every token of every rollout, so the signals are
superposed rather than partitioned. We further investigate whether the two signals
remain complementary when applied together, examining the training rollouts in
Section~\ref{ssec:mech}.

\begin{table*}[t]
\centering
\caption{Accuracy (\%) on nine benchmarks in three task families. Backbone: Qwen3.5-4B. V-OPD
abbreviates Vision-OPD. Post-trained
rows report mean $\pm$ SD over three seeds and share one training configuration and a $2048$-token
evaluation budget. Static-view distill matches the curriculum's expected view
exposure but removes its ordering. Best means are bold, and family averages are reported in the text.}
\label{tab:main}
\setlength{\tabcolsep}{1.5pt}
\renewcommand{\arraystretch}{1}
\begin{tabular}{lcccccccccc}
\toprule
 & \multicolumn{4}{c}{Perception \& spatial understanding} & \multicolumn{2}{c}{Visual math}
 & \multicolumn{3}{c}{General \& diagram QA} & \\
\cmidrule(lr){2-5}\cmidrule(lr){6-7}\cmidrule(lr){8-10}
Method & ZoomB. & HR-4K & HR-8K & CV-B.
       & MVista & MVerse
       & MMStar & MMMU & AI2D & \textbf{All} \\
\midrule
Qwen3.5-4B                    & 48.76 & 83.88 & 80.62 & 88.10 & 78.50 & 67.61 & 73.07 & 66.00 & 80.73 & 74.14 \\
GRPO & \result{57.51}{.59} & \result{\textbf{84.63}}{.63} & \result{80.25}{.38} & \result{87.81}{.16} & \result{76.70}{.60} & \result{63.10}{.19} & \result{72.02}{.57} & \result{63.59}{.61} & \result{85.11}{.31} & \result{74.53}{.30} \\
\midrule
V-OPD~\cite{visionopd} & \result{57.99}{.71} & \result{81.92}{.81} & \result{78.92}{.44} & \result{86.47}{.23} & \result{77.37}{.75} & \result{68.02}{.28} & \result{75.13}{.70} & \result{67.85}{.72} & \result{85.82}{.42} & \result{75.50}{.31} \\
Static-view distill & \result{56.96}{.60} & \result{81.83}{.69} & \result{79.79}{.51} & \result{85.97}{.19} & \result{78.20}{.80} & \result{69.35}{.27} & \result{75.07}{.64} & \result{67.78}{.78} & \result{86.03}{.37} & \result{75.67}{.28} \\
PVD-distill w/o RA & \result{57.71}{.65} & \result{83.42}{.56} & \result{80.79}{.31} & \result{86.57}{.18} & \result{78.73}{.55} & \result{70.95}{.24} & \result{75.33}{.67} & \result{67.22}{.67} & \result{86.29}{.28} & \result{76.34}{.29} \\
PVD-distill & \result{57.40}{.83} & \result{83.75}{.75} & \result{81.00}{.50} & \result{87.50}{.21} & \result{79.73}{.85} & \result{70.71}{.18} & \result{75.53}{.53} & \result{67.04}{.83} & \result{87.32}{.44} & \result{76.66}{.30} \\
\midrule
V-OPD w/ GRPO & \result{\textbf{58.26}}{.53} & \result{83.00}{.88} & \result{81.13}{.45} & \result{87.41}{.15} & \result{79.30}{.70} & \result{70.87}{.29} & \result{75.16}{.77} & \result{67.22}{.56} & \result{86.22}{.34} & \result{76.51}{.30} \\
\textbf{PVD-full} & \result{57.16}{.78} & \result{84.13}{.70} & \result{\textbf{81.21}}{.56} & \result{\textbf{88.32}}{.24} & \result{\textbf{81.13}}{.65} & \result{\textbf{72.00}}{.23} & \result{\textbf{76.38}}{.60} & \result{\textbf{69.33}}{.73} & \result{\textbf{87.94}}{.39} & \result{\textbf{77.51}}{.26} \\
\bottomrule
\end{tabular}
\end{table*}

\section{EXPERIMENTS}
\label{sec:exp}

\subsection{Setup}
\label{ssec:setup}

\noindent\textbf{Training.} All post-training runs are initialized from the same
Qwen3.5-4B checkpoint~\cite{qwen35} and trained for $65$ steps on the public
Vision-OPD-6K set~\cite{visionopd}. We use $96$ prompts per step with $8$ rollouts
each, learning rate $2\!\times\!10^{-6}$ with $10$ warmup steps, prompt and response budgets of
$8192$ and $1024$ tokens. Training and evaluation ran on 2 servers with $8$ NVIDIA A800 GPUs each,
using FSDP and vLLM rollouts~\cite{verl}. The teacher is an exponential-moving average of the policy,
$\theta_T \leftarrow (1-\alpha)\,\theta_T + \alpha\,\theta$ with $\alpha = 0.05$ applied every
optimizer step, and $K = 100$ in \eqref{eq:jsd}.
All post-trained rows of Table~\ref{tab:main} share this configuration, and each
ablation pair isolates one component (scheduling, then RA).

\noindent\textbf{Training data.} Vision-OPD-6K provides region-grounded perception
questions over high-resolution natural photographs, $6{,}240$ examples, $99.98\%$ four-option
multiple choice. Over a random sample of $400$,
the teacher's crop covers a median $7.10\%$ of the image area (IQR $1.90$--$20.30\%$), so
cropping presents the relevant evidence at a much larger effective scale than the full image.
On all examples the first view $v_1$ is the teacher's crop, so the $\delta_T = 0$ endpoint
is realized exactly and $g_t(v)$ reduces to a parametric teacher--student advantage under
identical visual inputs.

\noindent\textbf{Hyperparameters and why.} For \eqref{eq:centered-gap}--\eqref{eq:w} we use $\lambda = 0.5$,
$g_{\max} = 2$ and $[w_{\min}, w_{\max}] = [0.5, 2]$, bounding the ratio between the most
and least weighted token at $4$ so no single token dominates. The stage boundaries are
$\tau_1 = 0.2$, $\tau_2 = 0.6$ with probabilities $(0.7, 0.2, 0.1)$, $(0.3, 0.4, 0.3)$,
$(0.0, 0.2, 0.8)$ over (crop, padded crop, full image). The first boundary sits just after the
$10$-step warmup, so the crop-heavy stage covers the least stable part of training, and the
last stage spans $40\%$ of the run to let the policy adapt to the deployment view. Every stage keeps
mass on at least two views, so exposure to later views increases in increments rather than by a hard
switch. We set
$\beta = 1$ in \eqref{eq:joint}.

\noindent\textbf{Evaluation.} We report accuracy on benchmarks across three families:
perception and spatial understanding (ZoomBench~\cite{zoombench}, HR-Bench 4K and
8K~\cite{hrbench}, CV-Bench~\cite{cvbench}), visual mathematical reasoning
(MathVista~\cite{mathvista}, MathVerse~\cite{mathverse}), and general multimodal and diagram QA
(MMStar~\cite{mmstar}, MMMU~\cite{mmmu}, AI2D~\cite{ai2d}). Every row uses the same
protocol, namely greedy decoding with a $2048$-token budget, rule-based answer extraction, and a
single locally served Qwen3-30B-A3B-Instruct-2507 judge~\cite{qwen3} at temperature $0$ with one fixed prompt per
benchmark, scoring the extracted answer against the official answer key. The judge only maps a
free-form response onto that key. It never sees the model identity or competing predictions, and the
same judge is applied to every row.
The all-benchmark average weights the nine benchmarks equally.

\noindent\textbf{Compared systems.} \emph{Vision-OPD} (V-OPD)~\cite{visionopd} uses no task reward.
\emph{GRPO} uses task reward alone, and \emph{V-OPD w/ GRPO} adds the
same reward to that baseline, giving a reward-matched control. \emph{Static-view distill} samples the
same three views throughout with the time-averaged curriculum probabilities
$(0.26, 0.28, 0.46)$, matching expected exposure but removing the ordering.
\emph{PVD-distill w/o RA} restores the schedule, and \emph{PVD-distill} additionally uses the
regional-advantage token weighting of \eqref{eq:w}. These three are reward-free. \emph{PVD-full}
combines schedule, RA and reward.

\subsection{Main results}

PVD-full reaches $77.51\pm0.26$ overall in Table~\ref{tab:main}, improving on V-OPD by $2.01$ points
and on its reward-matched counterpart V-OPD with GRPO by $1.00$ point. The improvement over both holds
on all three task families, and PVD-full obtains the best mean on $7$ of the $9$ individual
benchmarks.
Relative to V-OPD, PVD-full improves $8$ of $9$ benchmarks, with family-level gains of $+1.38$
(perception), $+3.87$ (visual math) and $+1.62$ (general QA). Per-benchmark gains reach
MathVerse $+3.98$, MathVista $+3.76$, HR-8K $+2.29$ and AI2D $+2.12$. Against the reward-matched
control, PVD-full is better on $8$ of $9$ benchmarks. Removing reward from both sides preserves the
ordering. PVD-distill
improves $7$ of $9$ benchmarks and gains $1.16$ points over V-OPD. PVD-full also has the lowest
overall SD among post-trained methods ($0.26$ versus $0.28$--$0.31$).

\subsection{Ablation: schedule, exposure and token allocation}
\label{ssec:ablation}

\noindent\textbf{Progressive scheduling improves over exposure-matched static mixing.} Static-view
distill sees each view as often as the curriculum does in expectation and differs from PVD-distill
without RA only in \emph{when} each view is sampled. It reaches $75.67\pm0.28$, only $0.17$ points above
V-OPD, whereas the progressive schedule reaches $76.34\pm0.29$, $0.84$ points above V-OPD. This
$0.67$-point margin over the static mixture accounts for approximately $80\%$ of the schedule's
$0.84$-point gain over V-OPD. Family-level results sharpen this contrast. Static mixing captures much
of the visual-math gain ($+1.08$)
but loses perception ($-0.19$), which the schedule recovers ($+0.99$ over the static mixture). Crop
exposure alone is therefore insufficient to explain the perception gains. This control measures the
full stagewise schedule, including its stronger emphasis on the full image near the end, rather than
an isolated easy-to-hard ordering effect.

\noindent\textbf{Token allocation is a complementary refinement.} Adding RA to the progressive
objective raises the average from $76.34$ to $76.66\pm0.30$, so roughly $72\%$ of the $1.16$-point reward-free
improvement over V-OPD comes from the view curriculum. RA improves all three family averages and the overall accuracy on each of the three
matched seeds ($+0.33$, $+0.31$, $+0.34$). These gains span only $0.03$ points across 3 seeds, supporting a small
but repeatable effect. As
\eqref{eq:allocation} suggests, RA weighting refines a transfer path already effective with
uniform supervision. RA thus provides a modest but consistent additional gain over the
progressive-view curriculum, which remains the primary driver of the reward-free improvement.

\subsection{Progressive view sampling shapes token supervision}

\begin{table}[tb]
\centering
\caption{Stage diagnostics from the reward-free run. We report the mean gap $\langle g\rangle$, the
fraction of high-gap tokens with $g_t>0.10$, the mean unweighted JSD $\langle \ell\rangle$, and the
min--max range of normalized RA weights.}
\label{tab:sched}
\setlength{\tabcolsep}{3.2pt}
\renewcommand{\arraystretch}{1}
\begin{tabular}{lcccc}
\toprule
Stage & $\langle g\rangle$ ($10^{-2}$) & high-gap & $\langle \ell\rangle$ ($10^{-3}$) & weights \\
\midrule
1 & 2.4 & 5.9\% & 4.4 & $[0.97,1.88]$ \\
2 & 5.2 & 12.5\% & 12.2 & $[0.90,1.98]$ \\
3 & 4.6 & 12.0\% & 10.2 & $[0.90,1.98]$ \\
\bottomrule
\end{tabular}
\end{table}

Table~\ref{tab:sched} tracks the teacher--student gap and divergence as the schedule shifts toward
later, more deployment-aligned view mixtures. The mean privileged-view gap is $2.4$ in the crop-heavy
first stage and $5.2$ and $4.6$ in Stages 2 and 3, in units of $10^{-2}$. Likewise, the high-gap token
fraction changes from $5.9\%$ to $12.5\%$ and $12.0\%$, while the unweighted JSD of
\eqref{eq:jsd} changes from $4.4$ to $12.2$ and $10.2$ in units of $10^{-3}$. The later-stage
mixtures, which include the grid-aligned padded crop and increasing full-image exposure, are therefore
associated with a harder surrogate than the first stage. The RA weights widen from $[0.97,1.88]$ to
$[0.90,1.98]$, making token-level emphasis more selective under the same weighting rule. The lower
gap in Stage 3 than Stage 2 is consistent with reduced teacher--student disagreement as training progresses.

\subsection{Complementarity with task reward}
\label{ssec:mech}

Adding task reward raises PVD-distill by $0.85$ points to
$77.51\pm0.26$ and raises V-OPD by $1.01$ points. Reward-only GRPO and reward-free PVD-distill
reach $74.53$ and $76.66$, respectively, both below PVD-full, so the two signals remain complementary.
\begin{table}[tb]
\centering
\caption{Rollout-group composition averaged over the last $10$ training steps. Only
disagreeing groups produce a GRPO gradient.}
\label{tab:rollout}
\setlength{\tabcolsep}{3pt}
\renewcommand{\arraystretch}{1}
\begin{tabular}{lcccc}
\toprule
Objective & all correct & all wrong & disagree & zero PG grad. \\
\midrule
GRPO & 43.1\% & 23.8\% & 33.1\% & 66.9\% \\
V-OPD w/ GRPO & 42.4\% & 13.5\% & 44.1\% & 55.9\% \\
PVD-full & 41.0\% & 12.7\% & 46.3\% & 53.7\% \\
\bottomrule
\end{tabular}
\end{table}
Table~\ref{tab:rollout} examines the rollout groups that supply the GRPO signal. Each prompt
produces $8$ rollouts, and the group-relative advantage is zero when all are correct or all are
wrong. Reward-only GRPO leaves $66.9\%$ of
groups with zero gradient, including $23.8\%$ in which every rollout is wrong. Adding distillation
cuts the all-wrong fraction to $13.5\%$ for V-OPD with GRPO and $12.7\%$ for PVD-full, raising
the reward-gradient-producing share to $44.1\%$ and $46.3\%$. The dense per-token distillation loss built from
\eqref{eq:jsd} does not require a correct rollout, providing supervision where reward supplies no
within-group gradient; the jointly trained policies also yield more groups with nonzero reward gradients.

\subsection{Generalization beyond the training format}

Training uses only region-grounded perception questions over natural photographs, yet the gains are
larger on visual math and general QA than on the perception family that the training data resemble.
The pattern holds against the reward-matched V-OPD with GRPO, with gains of $+0.26$ on
perception, $+1.48$ on visual math and $+1.68$ on general QA. The gains therefore extend to questions
and imagery beyond the training format, consistent with transfer of improved fine-detail sensitivity.

\section{CONCLUSION}

We presented PVD, coupling a stagewise regional-to-global view schedule through a token-grid-aligned
padded crop with teacher-gap-based token reweighting. The padded crop preserves regional content while
aligning token geometry, and the schedule accounts for most of the reward-free gain. Combined with task reward, PVD-full
reaches $77.51\pm0.26$, exceeding reward-free and reward-matched baselines by $2.01$ and $1.00$
points, respectively. Deployment requires neither region annotations nor inference-time search.

% ---- page 5: ethics, acknowledgement and references only (ICASSP rule) ----
\clearpage
\section{COMPLIANCE WITH ETHICAL\\ STANDARDS}

This work did not involve human participants or animals, and no ethical approval was required.
No new data were collected, and all experiments use publicly released model weights, a publicly
released training set and publicly released multimodal evaluation benchmarks intended for research
use.

\section{ACKNOWLEDGMENT}

This work was conducted at Baidu Inc. ChatGPT was used for language editing and figure refinement of the manuscript
and limited assistance in debugging experimental code. All AI-assisted edits and code changes
were reviewed and validated by the authors, who take full responsibility for the content of this
work. The authors declare no relevant financial or non-financial interests.

\bibliographystyle{IEEEbib}
\bibliography{refs}

\end{document}